# Review of Feed-forward 3D Reconstruction: From DUSt3R to VGGT

Wei Zhang[1,2], Yihang Wu[1], Songhua Li[1], Wenjie Ma[1], Xin Ma[1], Qiang Li[1], Qi Wang[1,*]

1. *School of Artificial Intelligence, Optics and Electronics(iOPEN), Northwestern Polytechnical University, Xi'an 710072, China; 2. School of Computer Science, Northwestern Polytechnical University, Xi'an 710072, China*

**Abstract**

3D reconstruction, which aims to recover the dense three-dimensional structure of a scene, is a cornerstone technology for numerous applications, including augmented/virtual reality, autonomous driving, and robotics. While traditional pipelines like Structure from Motion (SfM) and Multi-View Stereo (MVS) achieve high precision through iterative optimization, they are limited by complex workflows, high computational cost, and poor robustness in challenging scenarios like texture-less regions. Recently, deep learning has catalyzed a paradigm shift in 3D reconstruction. A new family of models, exemplified by DUSt3R, has pioneered a feed-forward approach. These models employ a unified deep network to jointly infer camera poses and dense geometry directly from an Unconstrained set of images in a single forward pass. This survey provides a systematic review of this emerging domain. We begin by dissecting the technical framework of these feed-forward models, including their Transformer-based correspondence modeling, joint pose and geometry regression mechanisms, and strategies for scaling from two-view to multi-view scenarios. To highlight the disruptive nature of this new paradigm, we contrast it with both traditional pipelines and earlier learning-based methods like MVSNet. Furthermore, we provide an overview of relevant datasets and evaluation metrics. Finally, we discuss the technology's broad application prospects and identify key future challenges and opportunities, such as model accuracy and scalability, and handling dynamic scenes.

**Keywords**：3D Reconstruction, Multi-View Stereo, Deep Learning, Feed-forward Model.

## 1. Introduction

Inferring dense 3D geometry from a collection of 2D images is a fundamental problem in computer vision, with far-reaching applications in cartography, robotics, augmented/virtual reality, and cultural heritage preservation. For decades, the field has been dominated by a traditional, geometry-based methodology. This paradigm typically consists of two sequential stages: Structure from Motion (SfM) and Multi-View Stereo (MVS). SfM begins by extracting and matching sparse local features (e.g., SIFT [1]) across images, followed by an iterative Bundle Adjustment (BA) process to jointly optimize camera poses and a sparse 3D point cloud. The open-source toolbox COLMAP [2] stands as a prime example of this pipeline, establishing itself as a cornerstone for both academia and industry due to its high accuracy and completeness. Subsequently, MVS leverages these known camera poses to establish dense pixel-wise correspondences, estimate a per-pixel depth, and ultimately fuse these into a dense point cloud or mesh model. Despite the remarkable accuracy of this classic pipeline, it faces systemic challenges in terms of robustness, efficiency, and ease of use, stemming from its inherent iterative nature. The sequential workflow leads to error propagation across stages, rendering it particularly susceptible to challenging scenarios such as texture-less regions, wide baselines, or non-Lambertian surfaces. Furthermore, its high computational complexity and reliance on domain expertise often limit its potential in real-time and large-scale applications.

The advent of deep learning initially manifested as innovations targeting specific components within the traditional pipeline. For instance, SuperPoint [3] and SuperGlue [4] enhanced the robustness of feature extraction and matching. Concurrently, deep networks, exemplified by MVSNet [5] and its successors (e.g., CasMVSNet [6],

---

[*]: Corresponding author.

PatchmatchNet [7]), replaced the conventional MVS step with a learnable framework, achieving higher-quality dense reconstruction given pre-computed, accurate poses. However, these "component-wise" or "modular" improvements, while boosting local performance, did not fundamentally alter the iterative and often fragile nature of the SfM stage. The core bottleneck of the system persisted. This stagnation prompted a fundamental question: Is it possible to design a unified, end-to-end model that bypasses the intricate iterative optimization and directly infers globally consistent camera poses and dense geometry from an Unconstrained set of images in a single feed-forward pass?

A recent wave of research, pioneered by the seminal work of DUSt3R [8], has provided an affirmative answer to this question, thereby giving rise to a new research paradigm termed Feed-forward 3D Reconstruction. The central tenet of these methods is to internalize the complex geometric reasoning and optimization of the traditional pipeline into a single, powerful deep network—typically based on a Transformer [9] architecture—effectively "distilling" the entire SfM+MVS workflow. This marks a paradigm shift in 3D reconstruction from "iterative optimization" to "end-to-end inference." Since the debut of DUSt3R[8], the field has witnessed an explosion of related research in a remarkably short period. From works enhancing core correspondence quality (e.g., MASt3R [10], VGGT [11]) and addressing multi-view consistency (e.g., Align3R [12], Pow3R [13]), to those targeting specific applications like real-time SLAM (SLAM3R [14]), autonomous driving (Driv3R [15]), and visual relocalization (Reloc3r [16]), a comprehensive technological ecosystem is rapidly forming (see Figure 1). While each work has a distinct focus, they share a unified technical core of joint pose and geometry inference from learned dense correspondences. This signals that the research direction has achieved sufficient depth and breadth to warrant a systematic review.

This survey aims to fill this critical gap. We will systematically contrast the traditional pipeline with the feed-forward paradigm to elucidate the latter's fundamental innovations. We will then dissect the core technical architecture of feed-forward models, including their Transformer-based correspondence modeling, joint inference mechanisms, and multi-view extension strategies. We further discuss how this paradigm reshapes the 3D reconstruction landscape and explore its broad application prospects. Finally, we identify key challenges and opportunities for future research, including avenues for improving accuracy and scalability, handling dynamic scenes, and integrating with neural implicit representations. Through this work, we hope to provide researchers with a clear roadmap, fostering further innovation and advancing the field.

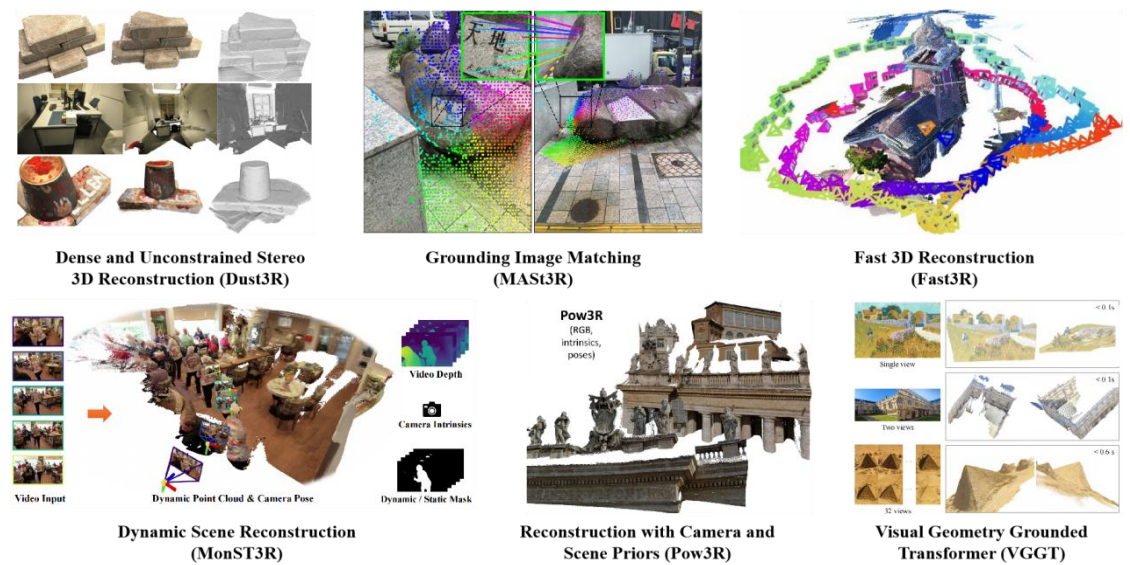

**Fig. 1. DUSt3R [8] and subsequent improvements.**

## 2. Framework of Feed-forward Reconstruction

The emergence of feed-forward 3D reconstruction models is predicated on the creation of a highly integrated deep network. This network internalizes the discrete, iterative steps of traditional methods—feature extraction, matching, pose estimation, and depth reconstruction—into a single, end-to-end forward inference process. While specific implementations vary, these models generally adhere to a coherent architectural logic. The process begins with a powerful encoder extracting deep image features. Next, a Transformer-based global matching module establishes dense, pixel-wise correspondences. Finally, a joint decoding head simultaneously infers the relative camera poses and the scene's 3D geometry from these correspondences. This section dissects this core technical framework, breaking it down into its three fundamental pillars: (1) learning robust dense correspondences, (2) joint inference of geometry and pose, and (3) scaling from two-view to multi-view scenarios. Table 1 provides a comprehensive summary of key models in this domain, highlighting their core contributions and architectural innovations.

**Learning Robust Dense Correspondences:** A primary bottleneck of traditional methods is their reliance on sparse, hand-crafted local features (e.g., SIFT [1]), which are prone to failure in texture-poor regions or under significant viewpoint changes. To overcome this limitation, feed-forward models draw inspiration from LoFTR [17] and its successors, employing the Transformer architecture for feature matching. Typically, input images are first processed by a Convolutional Neural Network (CNN) or a Vision in Transformer (ViT) [18] backbone to extract multi-scale feature maps. These maps are then flattened into sequences of tokens and fed into a Transformer encoder, which consists of stacked self-attention and cross-attention layers. The self-attention layers enhance the contextual awareness of features within each image, allowing every pixel's feature to "see" the entire image. The cross-attention layers, in contrast, facilitate information exchange between tokens from different images, enabling the computation of pixel-wise matching relationships within a global receptive field. The seminal work of DUSt3R [8] was the first to fully unify this process with pose and depth estimation. Subsequent research has explored various avenues to further enhance correspondence quality. For instance, MASt3R [10] and VGGT [11] focus on improving matching accuracy in wide-baseline and varying-resolution scenarios through more sophisticated multi-scale feature fusion strategies. SPANN3R [19] investigates sparse attention mechanisms to mitigate the high computational cost of Transformers on high-resolution images while preserving their global matching capabilities. Collectively, these models have shifted the output of the matching stage from discrete "inlier/outlier" pairs to a dense, probabilistic confidence map, providing a much richer signal for downstream geometric inference.

**Joint Inference of Geometry and Pose:** In the traditional pipeline, this stage would necessitate robust sampling algorithms like RANSAC to iteratively solve for the essential or fundamental matrix from noisy correspondences. Feed-forward models, however, introduce a novel regression head that takes the entire dense correspondence map as input to directly decode geometric information. The implementation in DUSt3R [8] is particularly elegant: instead of directly regressing the rotation matrix R and translation vector t, it predicts the 3D coordinates (i.e., a point cloud) for each pixel in the first image, expressed in a normalized camera coordinate system. A differentiable Kabsch or Umeyama algorithm layer then computes the optimal rigid transformation that aligns this predicted point cloud with its counterpart projected from the second image via the established correspondences. This transformation itself yields the desired relative pose (R, t), while the predicted 3D coordinates directly provide the dense scene geometry. This "pose-from-alignment" approach gracefully couples the estimation of pose and geometry, enabling them to mutually supervise each other. Building on this foundation, various models have explored diverse decoding strategies. PE3R [47], for example, focuses on more directly regressing the parameters of the essential matrix from correspondences, incorporating loss functions to enforce its algebraic properties. MonST3R [20] demonstrates the framework's flexibility by effectively fusing priors from a monocular depth estimation network during the decoding phase. This fusion resolves the inherent scale ambiguity of two-view reconstruction, enabling metrically scaled results. Meanwhile, PreF3R [21] explores a hybrid paradigm, using the feed-forward model to provide a high-quality initial pose, which is then refined by a lightweight iterative optimization module. This approach aims to combine the robustness of feed-forward models with the precision of traditional optimization.

**Scaling from Two-View to Multi-View Scenarios:** Given the prohibitive computational complexity of directly processing N images, the vast majority of models adopt a two-stage strategy: first, run the two-view model on all or a subset of image pairs, and then globally aggregate the resulting pairwise estimates. Align3R [12] made pioneering contributions in this direction by proposing a confidence-weighted pose graph optimization algorithm. This method robustly solves for globally consistent camera poses from a large set of pairwise relative poses, which may contain significant errors. Pow3R [13] advanced this concept by not only optimizing the pose graph but also incorporating all two-view point cloud fragments into the global optimization objective. By jointly optimizing camera poses and point cloud alignment, it achieves a higher level of geometric consistency. REGIST3R [22] offers a similar framework for robust registration in complex scenes. Beyond this general-purpose aggregation, specialized sequential processing methods have been developed for specific data modalities. SLAM3R [14] and Driv3R [15] successfully adapt the paradigm to real-time SLAM and autonomous driving by introducing key-frame management and loop closure mechanisms, enabling efficient incremental reconstruction and localization from video streams. Nevertheless, the pursuit of a truly end-to-end multi-view model continues. Cutting-edge works like MV-DUSt3R+ [23] are beginning to design Transformer architectures capable of directly ingesting N>2 images. By modifying the attention mechanism to enable simultaneous information exchange across multiple views, these models theoretically bypass the error accumulation associated with pairwise aggregation, though this places significantly higher demands on GPU memory and computation. Furthermore, some works like Easi3R [24] introduce an iterative refinement concept within a single feed-forward network. By cyclically updating correspondences and geometry between network layers, they aim to achieve higher precision in a single forward pass, cleverly merging inference with optimization. These diverse extension strategies, alongside ongoing explorations in efficiency (Fast3R [25], CUT3R [26]), specific applications (Reloc3r [16] for relocalization, AerialMegaDepth [28] for aerial imagery), and other areas, collectively demonstrate the immense potential of this framework as a flexible and powerful foundation for 3D perception.

**Table 1 Comprehensive Summary of Key Models in Feed-forward Reconstruction.**

| Model | Contribution | Input | Output | Primary Application |
|---|---|---|---|---|
| DUSt3R [8] | Pioneering work that unifies relative pose estimation and dense reconstruction for two-view inputs. Introduces a confidence-weighted aggregation scheme. | Image pairs | Relative pose, Point cloud | General-purpose 3D Reconstruction |
| MASt3R [10] | Enhances the quality and robustness of multi-scale dense matching, serving as a powerful correspondence engine for challenging scenes. | Image pair | Dense correspondences, Relative pose | Wide-baseline matching & reconstruction |
| Align3R [12] | Introduces an efficient multi-view alignment algorithm to robustly aggregate pairwise estimates into | Set of pairwise results | Global poses | Large-scale scene reconstruction |

| | globally consistent camera poses. | | | |
|---|---|---|---|---|
| Pow3R [13] | Extends pose graph optimization by incorporating point cloud registration, enabling joint global optimization of both camera poses and geometry. | Image set | Global poses, Fused point cloud | High-consistency scene reconstruction |
| SLAM3R [14] | Adapts the feed-forward paradigm to real-time SLAM by introducing keyframe management and incremental update mechanisms. | Image sequence | Real-time camera trajectory, Map | Robotics, Navigation, AR |
| Fast3R [25] | Focuses on model efficiency, achieving faster reconstruction speeds through lightweight architectural design and knowledge distillation. | Image pair | Relative pose, Point cloud | Resource-constrained devices |
| Driv3R [15] | Specifically designed for autonomous driving, leveraging Bird's-Eye-View (BEV) representations and temporal information. | Video sequence | Vehicle trajectory, Dynamic objects | Autonomous driving perception |
| MonST3R [20] | Integrates monocular depth priors to resolve scale ambiguity and enhance the metric-scale accuracy of the reconstruction. | Image pair | Metric-scale pose, Point cloud | Monocular scale recovery |
| MV-DUSt3R+ [23] | Pioneers a true multi-view architecture that directly processes a set of N>2 images, bypassing the need for pairwise aggregation. | Image set (N>2) | Global poses, Point cloud | End-to-end multi-view reconstruction |
| Reloc3r [16] | Applies the feed-forward paradigm to the visual relocalization task, estimating the pose of a query image with respect to a pre-existing map. | Query image + Map | Camera pose | AR, Robot relocalization |

**3. Comparison with Classical MVS**

The advent of feed-forward models represents not merely an algorithmic innovation, but a disruptive reconfiguration of the entire 3D reconstruction pipeline. This transformation can be understood through three fundamental shifts: a move from iterative pipelines to holistic inference, a philosophical change from explicit geometric constraints to implicit data-driven priors, and a relocation of developmental and computational bottlenecks. A detailed comparison is shown in Table 2.

**Unification of the Reconstruction Workflow:** First, feed-forward models fundamentally redefine the 3D reconstruction workflow, transforming it from a multi-stage, sequential, and iterative process into a single, parallelizable inference task. The conventional pipeline, exemplified by systems like COLMAP [2], consists of a complex and fragile cascade of algorithms. This process commences with sparse feature extraction (e.g., SIFT [1]), proceeds to exhaustive or sequential feature matching, applies geometric verification for outlier rejection (e.g., RANSAC), and culminates in an incremental framework of repeated triangulation and Bundle Adjustment (BA). The resulting sparse model then serves as input to a Multi-View Stereo (MVS) module (e.g., MVSNet [5], CasMVSNet [6]) for densification. Each stage in this pipeline is critically dependent on the output quality of the preceding one, where a failure can compromise the entire reconstruction process. In contrast, the emergent paradigm, represented by models such as DUSt3R [8], encapsulates this complex procedure within a single deep neural network. This allows for the direct processing of an unconstrained set of images to produce globally consistent camera poses and dense geometry in a single forward pass. Even for two-stage approaches—such as employing MASt3R [10] for high-quality pairwise estimation followed by global alignment with Align3R [12] or Pow3R [13]—the workflow is reduced to a concatenation of two inference steps, thereby avoiding the non-deterministic and computationally intensive iterative loops characteristic of traditional SfM. This marks a paradigm shift from a multi-stage algorithmic pipeline to a unified, end-to-end inference framework, which significantly reduces system complexity and enhances overall robustness.

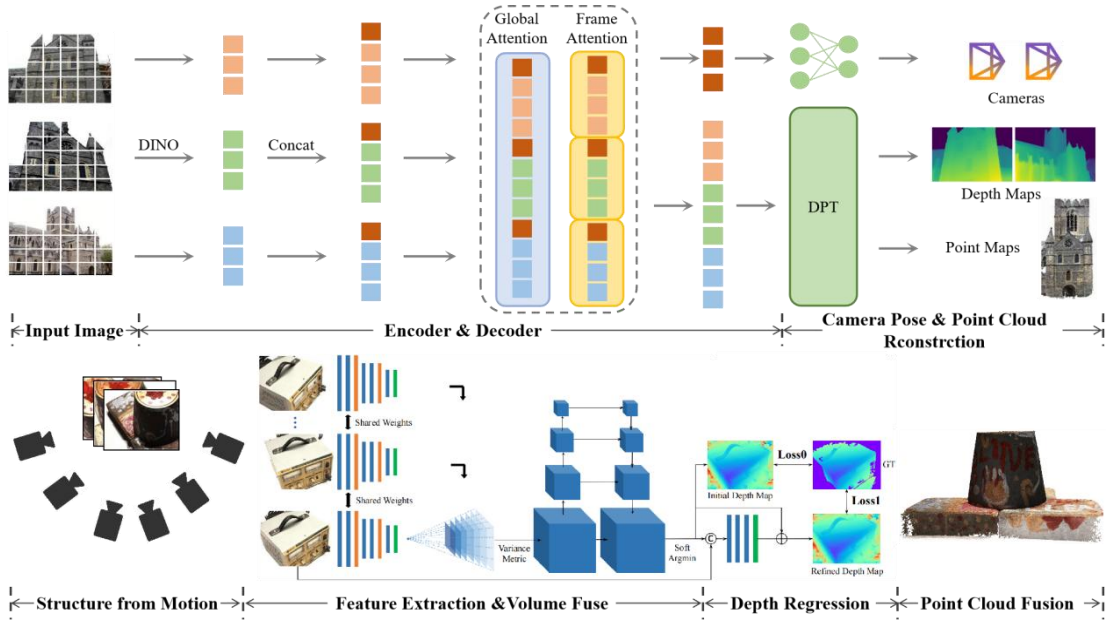

**Fig. 2. Comparison with Classical MVS. The upper panel illustrates the VGGT framework. It takes multi-view images as input and directly regresses camera poses and dense depth maps through a unified encoder-decoder network. The lower panel depicts the conventional, multi-stage MVS pipeline. This classical approach involves a sequence of discrete steps: first, SfM is used to estimate camera poses and a sparse point cloud; next, a separate depth regression module is employed for each view; finally, these individual depth maps are fused to generate a dense 3D point cloud.**

**From Explicit Geometric Models to Implicit Data-Driven Priors:** Second, this reconfiguration is rooted in a fundamental shift in modeling principles: from a reliance on explicit geometric constraints to the utilization of implicit, data-driven priors. The foundation of conventional methods rests upon rigorous mathematical models, such as epipolar geometry, the Perspective-n-Point (PnP) problem, and the Jacobian matrix of Bundle Adjustment. These methods formulate 3D reconstruction as an optimization problem constrained by well-defined geometric laws. However, their performance degrades significantly when input data violates the underlying assumptions of these models, as is common with textureless surfaces or severe illumination changes. Feed-forward models adopt a different approach, shifting the core dependency from explicit formulations to knowledge learned from data. By training on large-scale datasets like MegaDepth [27], these architectures learn powerful priors about real-world scene structure. The network implicitly encodes this knowledge within its weights, learning statistical regularities such as the planarity of surfaces, the typical distance of environmental elements like the sky, and the continuity of object shapes. Consequently, even in scenarios where conventional methods fail due to a lack of salient feature points, models like DUSt3R [8] or MASt3R [10] can infer plausible correspondences and geometry by leveraging these learned scene priors. This transition from model-based systems with explicit rules to data-driven inference is the primary reason for their enhanced robustness in challenging scenes.

**Shift in Computational and Developmental Bottlenecks:** Finally, the reconfiguration of the technical pipeline results in a corresponding shift in developmental and computational bottlenecks. In the conventional framework, research primarily focused on designing superior feature descriptors, more robust sampling strategies (e.g., RANSAC variants), and more efficient non-linear optimizers (e.g., BA solvers). The system's performance was principally constrained by the CPU-intensive iterative optimization process, a procedure often requiring minutes to days of computation. Under the feed-forward paradigm, the emphasis on manual algorithm design is largely supplanted by the demands of large-scale model training. Consequently, the primary bottlenecks have relocated to new areas: 1) GPU Compute and Memory: Training and deploying large Transformer models require substantial GPU resources, which has motivated research into model compression and efficient attention mechanisms, as exemplified by works like Fast3R [25]. 2) Data Curation: Model performance is now highly contingent on the scale, diversity, and quality of training data. The acquisition of large-scale datasets with accurate ground-truth poses and dense depth remains a significant hurdle. 3) Generalization to Unseen Domains: Ensuring robust model performance on out-of-distribution data has become a critical area of investigation. This transition has consequently altered the requisite expertise for 3D vision researchers, demanding a shift from a focus on geometry and optimization towards neural network architecture design, self-supervised or weakly-supervised learning, and large-scale data management.

**Table 2 The Comparison of Classical Pipeline and Feed-forward Pipeline.**

| Aspect | Traditional Pipeline (e.g., COLMAP [2]) | Modular Deep Learning (e.g., MVSNet [5]) | Feed-forward Pipeline (e.g., DUSt3R [8]) |
|---|---|---|---|
| Workflow | Multi-stage, Sequential, Iterative | Modular, still Iterative | End-to-end, Feed-forward |
| Core Engine | Geometric Optimization (BA) | Learned Matcher + Geometric Opt. (BA) | End-to-End Network (Transformer) |
| Robustness | Sensitive to texture, baseline, init. | Improved matching, but fragile SfM core | High robustness via learned priors |
| Ease of Use | Requires expertise, complex tuning | Requires toolchain integration | "Black-box" model, direct inference |
| Computation Mode | Mixed CPU (BA) & GPU (MVS) | Mixed CPU (SfM) & GPU (MVS) | Fully GPU-intensive |
| Bottleneck | Geometric failures, optimization fragility | Iterative and fragile SfM core | GPU memory, Data scale & diversity |

**4. Dataset and metrics**

The emergence of feed-forward 3D reconstruction is closely tied to large-scale, diverse datasets. Unlike traditional methods based on explicit geometric principles, these deep learning models encode implicit geometric priors by training on vast data. Consequently, the scale, diversity, and accuracy of the training data directly govern a model's generalization capabilities and final performance. Prominent datasets used for training, such as Habitat [29], MegaDepth[27], ARKitScenes [30], CO3D-v2 [34], and Waymo [35], span diverse conditions, including indoor and outdoor environments, synthetic and real-world captures, and both static and dynamic scenes. This variety provides the necessary data foundation for training robust models equipped to handle real-world perceptual challenges.

**Table 3 Prominent Training Datasets in Feed-forward Model.**

| Dataset | Year | Size / Content | Category |
|---|---|---|---|
| Habitat (Matterport3D) [29] | 2019 | 1,000 scenes | Indoor |
| MegaDepth [27] | 2018 | 130K+ images, 200+ scenes | Outdoor |
| ARKitScenes [30] | 2021 | 5,048 RGB-D sequences from 1,661 scenes | Indoor |
| Static Scenes 3D [31] | 2016 | 35,000+ stereo image pairs | Outdoor |
| BlendedMVS [32] | 2020 | 17K+ images | Outdoor |
| ScanNet++ [33] | 2023 | 460 scenes, 280K DSLR images, 3.7M+ iPhone RGBD frames | Indoor |
| CO3D-v2 [34] | 2021 | 1.5 million frames from ~19,000 videos | Object-centric |
| Waymo [35] | 2020 | 1,150 scenes | Outdoor, Driving |
| DL3DV [36] | 2022 | 10,510 scenes | Outdoor |
| WildRGB-D [37] | 2024 | 13 long-term sequences, ~100K RGB-D frames | Indoor, Dynamic |
| HyperSim [38] | 2021 | 774 scenes (~100K images) from 461 distinct layouts | Indoor, Synthetic |
| Replica [39] | 2019 | 18 highly photorealistic 3D mesh reconstructions of indoor scenes | Indoor, Synthetic |
| MVS-Synth [40] | 2020 | 130 scenes with diverse styles | Outdoor / Object, Synthetic |
| Virtual KITTI [41] | 2016 | 50 high-resolution video sequences cloned from 5 KITTI scenes | Outdoor, Street-level/Driving |
| Aria Synthetic Environments [42] | 2023 | 100 digital twins of real-world indoor spaces | Indoor, Egocentric, Synthetic |
| Aria Digital Twin [42] | 2023 | 130 data sequences from 11 real-world scenes | Indoor, Egocentric |

The evaluation of feed-forward models in 3D vision is task-specific, employing distinct benchmarks and metrics to ensure rigorous assessment. For multi-view depth estimation, performance in predicting dense depth maps is benchmarked on datasets such as DTU [43], Tanks and Temples [44], ScanNet [48], and ETH3D [45]. Key metrics include Absolute Relative Error (*rel*), Inlier Ratios ($\tau$), and the mean distances for Accuracy (*Acc*) and Completeness (*Comp*) in *mm*, which collectively quantify the consistency between predicted and ground-truth depth. Similarly, the quality of 3D reconstruction is evaluated by assessing the geometric fidelity of the final fused model, primarily using the DTU [43] dataset. This evaluation relies on comparing the reconstructed point cloud to a ground-truth scan with three core metrics (in mm): Accuracy, the mean distance from the prediction to the ground-truth; Completeness, the mean distance from the ground-truth to the prediction; and the Overall Score, their average, which is equivalent to the Chamfer distance. The evaluation results for the two tasks are shown in Table 4. These standardized protocols facilitate fair model comparison and drive progress in the field.

**Table 4 Multi-view depth and 3D reconstruction evaluation on the DTU [43] Dataset.**

| Type | Model | Known GT camera | Multi-view depth evaluation | | | 3D reconstruction | | |
|---|---|---|---|---|---|---|---|---|
| | | | rel↓ | $\tau$↓ | time (s)↓ | Acc.↓ | Comp.↓ | Overall↓ |

|  | Model | | | | | | | |
|---|---|---|---|---|---|---|---|---|
| Traditional Model | Gipuma [49] | ✓ | / | / | / | 0.283 | 0.873 | 0.578 |
| | COLMAP [2][50] | ✓ | 0.7 | 96.5 | ~3 min | | | |
| | MVSNet [5] | ✓ | (1.8) | (86.0) | 0.07 | 0.396 | 0.527 | 0.462 |
| | MVSNet Inv. Depth [5] | ✓ | (1.8) | (86.7) | 0.32 | / | / | / |
| | CIDER [51] | ✓ | / | / | / | 0.417 | 0.437 | 0.427 |
| | PatchmatchNet [7] | ✓ | / | / | / | 0.427 | 0.377 | 0.417 |
| | GeoMVSNet [52] | ✓ | / | / | / | 0.331 | 0.259 | 0.295 |
| Feed-forward Model | DUSt3R[8] | ✗ | 3.3 | 69.9 | 0.05 | 2.667 | 0.805 | 1.741 |
| | VGGT [11] | ✗ | / | / | / | 0.389 | 0.374 | 0.382 |
| | MASt3R [10] | ✗ | / | / | / | 0.403 | 0.344 | 0.374 |
| | Test3R[53] | ✗ | 2.0 | 84.1 | / | / | / | / |
| | PE3R [47] | ✗ | 3.2 | 69.1 | / | / | / | / |
| | Spann3R [19] | ✗ | 3.5 | 65.2 | 0.32 | / | / | / |
| | MUSt3R [54] | ✗ | 4.6 | 63.1 | 0.19 | / | / | / |
| | Pow3R [13] | ✗ | 3.0 | 74.3 | / | 2.116 | 1.370 | 1.743 |

## 5. Applications and Future Challenges

The substantial robustness and operational simplicity of feed-forward reconstruction models have enabled a wide range of novel applications, particularly in unconstrained "in-the-wild" scenarios where traditional methods exhibit limitations. Their real-world impact is already evident across numerous domains. In consumer AR/VR, where real-time dense reconstruction is a significant objective, these models allow users to generate a 3D scan of a room in seconds using a smartphone—a task for which classical SfM/MVS pipelines are often computationally prohibitive or prone to failure. For robotics and autonomous systems, the capacity for low-latency pose and depth estimation significantly enhances system reliability, especially in complex environments. Consequently, robust SLAM (e.g., SLAM3R [14]) and localization (e.g., Reloc3r [16]) are now feasible under conditions that pose considerable challenges for conventional techniques. Furthermore, these models benefit fields such as emergency response and cultural heritage, where rapid, on-site 3D mapping from uncalibrated cameras is of critical value. A key factor underpinning this versatility is generalization. Trained on highly diverse datasets, models like DUSt3R[8] often exhibit zero-shot capabilities on novel scenes, indicating the acquisition of a powerful geometric prior that extends beyond the specific characteristics of their training data.

Despite these advances, several significant challenges remain unresolved. Scalability to large-scale environments constitutes a primary limitation. While models like VGGT[11] and Regist3R [22] demonstrate applicability to thousands of views, the quadratic complexity inherent in attention mechanisms imposes a fundamental bottleneck, precluding single-pass, city-scale reconstruction. The reconstruction of dynamic and non-rigid scenes, despite progress from works like MonST3R [20], remains suboptimal when confronted with complex, multi-object motion. Another critical and underexplored area is uncertainty quantification. For safety-critical applications such as robotics, principled methods for estimating the reliability of the reconstructed geometry are essential. This necessitates a transition from simplistic confidence maps to more sophisticated probabilistic frameworks, such as Bayesian or multi-hypothesis estimation. Finally, ensuring consistent performance on out-of-distribution (OOD) data and across diverse camera models remains an open research question, highlighting the need for even larger and more varied "3D foundation datasets."

Future research trajectories are oriented towards the development of versatile, multi-modal 3D foundation models, analogous to those in computer vision and natural language processing. A prominent research avenue involves hybrid architectures that integrate the complementary strengths of different paradigms: a feed-forward network for robust, rapid inference, coupled with a differentiable optimization layer for high-fidelity geometric refinement. This approach would merge the robustness of learned models with the precision of classical geometric optimization. Another significant frontier is the direct integration with neural rendering techniques. Future models may directly output implicit scene representations, such as Neural Radiance Fields (NeRFs) or 3D Gaussian Splatting grids, thereby facilitating real-time novel-view synthesis from sparse image inputs, a path explored by works like PreF3R [21] and MVSplat [46]. Ultimately, the synergy between 3D perception and natural language processing presents substantial opportunities for innovation. This could foster a new class of "geometrically-aware LLMs" capable not only of 3D reconstruction but also of semantic reasoning, spatial description, and interactive engagement with the reconstructed environment.

## 6. Conclusion

The field of 3D reconstruction stands at an exciting inflection point. Feed-forward networks, exemplified by DUSt3R [8] and its derivatives, are profoundly reshaping the technical landscape once dominated by multi-stage iterative optimization, championing a new philosophy of end-to-end inference. With their unprecedented robustness and simplicity, they are democratizing the capability for high-quality 3D reconstruction, moving it beyond the confines of expert users. This survey has systematically reviewed the origins, core techniques, model evolution, and transformative impact of this emerging paradigm on the reconstruction pipeline. We have argued that, despite ongoing challenges in areas such as accuracy and scalability, this direction has unequivocally ushered in a new era of 3D vision research. Future exploration will likely revolve around the trade-offs between precision and robustness, deeper integration with implicit neural representations, and extensions toward more complex and dynamic worlds. The immense potential of this paradigm promises the advent of a more intelligent, accessible, and ubiquitous era of 3D perception.

1050–1060.